# Immersive Human-in-the-Loop Control: Real-Time 3D Surface Meshing and Physics Simulation

Sait Akturk[†], Justin Valentine[†], Junaid Ahmad[†], Martin Jagersand[†]

*Abstract*— This paper introduces the TactiMesh Teleoperator Interface (TTI), a novel predictive visual and haptic system designed explicitly for human-in-the-loop robot control using a head-mounted display (HMD). By employing simultaneous localization and mapping (SLAM) in tandem with a space carving method (CARV), TTI creates a real-time 3D surface mesh of remote environments from an RGB camera mounted on a Barrett WAM arm. The generated mesh is integrated into a physics simulator, featuring a digital twin of the WAM robot arm to create a virtual environment. In this virtual environment, TTI provides haptic feedback directly in response to the operator's movements, eliminating the problem with delayed response from the haptic follower robot. Furthermore, texturing the 3D mesh with keyframes from SLAM allows the operator to control the viewpoint of their Head Mounted Display (HMD) independently of the arm-mounted robot camera, giving a better visual immersion and improving manipulation speed. Incorporating predictive visual and haptic feedback significantly improves tele-operation in applications such as search and rescue, inspection, and remote maintenance.

## I. INTRODUCTION

A central challenge in the field of robotics is the development of robots capable of navigating and interacting effectively in complex 3D environments in real-time. Practical approaches often involve human-in-the-loop operation, ranging from telemanipulation to semi-autonomous control. This approach finds particular applicability in high-risk areas [1], [2], medical assistance, and surgery [3]–[5], as well as disaster relief scenarios [6], [7].

Teleoperation in robotics faces significant challenges in real-world settings, with communication delays exceeding several seconds being one of the most critical issues affecting operator performance and task completion rates. [8], [9]. Additionally, the need for real-time visual and haptic feedback from remote environments demands low latency and high bandwidth networks. To address these challenges, we employ a real-time 3D surface mesh reconstruction algorithm for immersive visual predictive display and haptic surface interaction feedback.

Our approach involves using SLAM and the CARV algorithm with a monocular camera to generate a real-time surface mesh, preserving essential 3D geometric and visual details crucial for operator spatial awareness [10]. Leveraging the predictive visualization system proposed by Lovi et al. [11], which updates a 3D mesh surface model based on the operator's motion, we provide haptic feedback derived from the models, eliminating haptic delays between the operator and the follower robot.

[†]Department of Computing Science, University of Alberta, Edmonton AB., Canada, T6G 2E8. `{akturk, jvalenti, jahmad, mj7}@ualberta.ca`

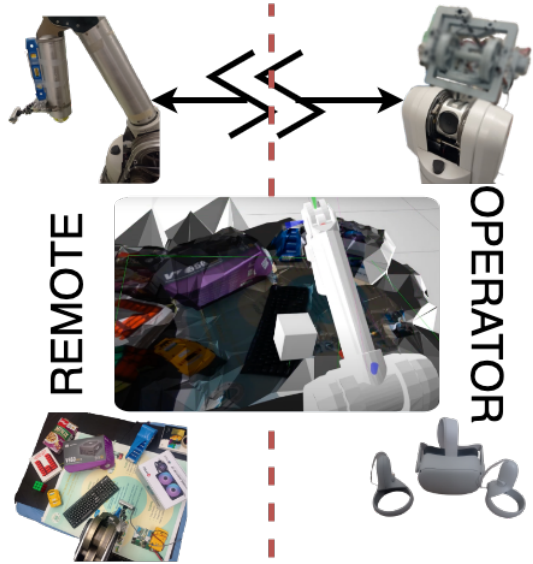

Fig. 1: Reconstructed surface mesh and texture of remote environment obtained by the monocular camera with a digital twin of remote WAM arm robot and Gazebo wide-angle camera sensor**(Cube)**.

In light of this, this paper further addresses a significant concern: the challenge of spatial awareness for operators managing remote teleoperated robotic systems. The ongoing need to assess the state of the robot arm and visualize the 3D representation of the distant environment can lead to mental load for operators.

To alleviate this issue, recent research has been focused on virtual reality (VR) and augmented reality (AR) immersive displays that provide 3D geometric information [12], [13]. One method for obtaining 3D geometric data involves using depth-sensing cameras. However, the dense point clouds generated by these devices are not ideal for teleoperation. Specifically, the high computational cost associated with transmitting and rendering these dense point clouds in real-time on a large scale poses a significant challenge [14]. On top of that, the absence of polygonal mesh connectivity and topological consistency leads to a loss in the geometric structure of the remote scene. As a result, the operator may struggle to fully comprehend or interact with the remote environment effectively [15].

To overcome these challenges, the research community put effort into real-time 3D reconstruction algorithms based on different 3D representations including voxels, point clouds, surfels and surface meshes. Compare to other 3D representation, surface meshes demand lower disk size, are easier to transfer and render in real-time, and provide complete 3D geometric information [16]. While machine learning-based approaches have advanced a wide range of computer vision and graphics tasks in recent years, the issue of surface mesh reconstruction from dense point clouds continues to present a formidable challenge for both deep learning-based and traditional methods [17], [18]. We propose using a semi-dense monocular CARV method to create a surface mesh of the remote environment in real-time, without excessive demands on computing power, memory, or network bandwidth [19].

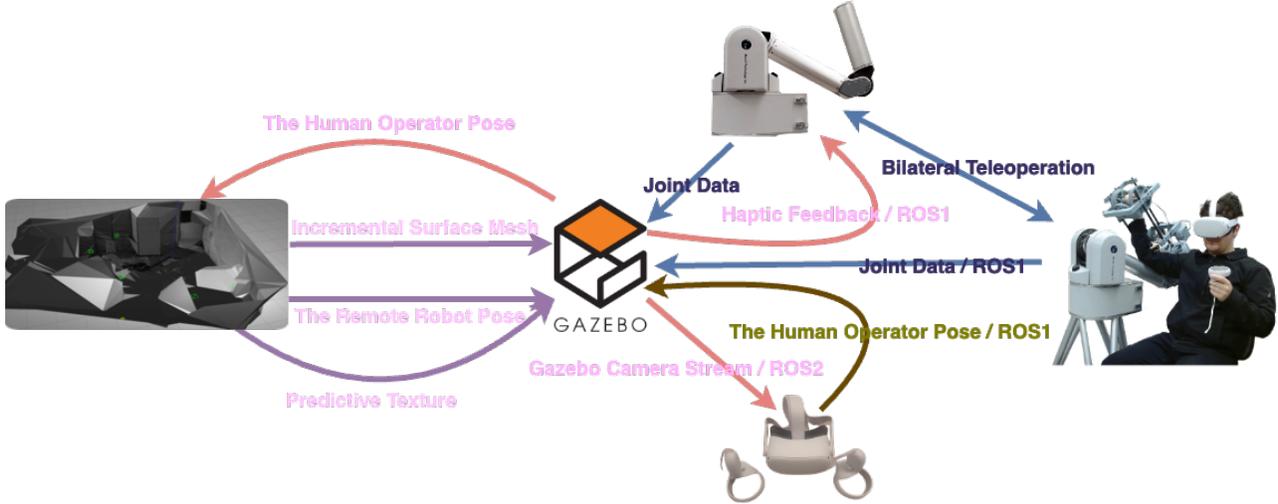

Fig. 2: Our real-time physics simulation-based immersive virtual world system. **(Left)** Incremental semi-dense CARV for surface mesh reconstruction, **(Bottom)** Immersive display with the controller, **(Top)** Remote WAM arm robot arm, **(Right)** Human controller with WAM arm robot and head-mounted Display, **(Middle)** Gazebo physics simulator to integrate digital twin of WAM arm robot, surface mesh and texture, Gazebo wide-angle camera sensor controlled by the human operator.

The primary contribution of this paper is to design a novel predictive system that offers haptic feedback to the human operator, thereby addressing and mitigating the delays typically associated with the haptic follower robot. The distinct components of our system include:

- **Incremental Surface Mesh:** We employ the semi-dense monocular CARV algorithm to produce an incremental surface mesh of the remote environment.
- **Physics Simulator:** The Gazebo physics simulator enables real-time visualization and interaction by modeling contact forces between the incremental surface mesh and the digital twin of the WAM robot arm.
- **Immersive VR Display:** This module employs a VR headset to control a pair of wide-angle stereo cameras within the Gazebo simulation environment, immersing the operator within the virtual environment.

## II. RELATED WORK

While robotics has been successful in repetitive tasks in known environments, the challenge of achieving robot autonomy and control in unstructured environments remains. Recent advances in control systems, computing, and artificial intelligence have opened up possibilities for addressing these challenges. However, factors such as the complexity of dynamical systems, the necessity for real-time processing, and limitations on power and computing resources make it difficult for robots to navigate and interact in unknown 3D environments [20].

In light of these challenges, human-in-the-loop teleoperation has surfaced as a compelling alternative, especially in mission-critical scenarios like search and rescue, hazardous work situations, medical surgeries, and telenursing. Supported by decades of research in robot teleoperation, human-in-the-loop systems alleviate the need for the robot to have a comprehensive grasp of its environment. Specifically, they enable human operators to plan and manipulate actions in 3D unknown environments, thereby offsetting the robot's limitations. However, the success of teleoperation is highly dependent on real-time visual and control feedback from the follower robot, which allows the operator to process information, make decisions, and execute actions effectively.

In the subsections that follow, we review related work in surface reconstruction, immersive display, and physics simulation. We examine the challenges in each area and discuss recent advances and approaches that have been proposed to overcome these challenges.

## A. Surface Reconstruction

Real-time 3D surface reconstruction is a crucial component of robot teleoperation. It enables the robot operator to have better spatial awareness and reduces mental load resulting modelling environment during control of the follower robot [21].

Numerous research efforts have been directed at capturing dense point clouds in remote environments using depth cameras to enhance robot teleoperation. However, using these dense point clouds for real-time visualization in teleoperation poses considerable challenges. These challenges relate to memory consumption, computational overhead, and network bandwidth. Together, these factors limit both the viability and scalability of remote scene representation [22]. An alternative to dense point cloud-based 3D scene reconstruction is the use of volumetric representation, which offer more efficient data representation and thereby make real-time reconstruction for robot teleoperation feasible [23].

Surface mesh reconstruction, featuring polygonal mesh connectivity, presents another viable option. It can represent remote scenes using either dense or sparse point cloud approaches. It offers the advantages of being easily updatable and transferable in real-time with low memory consumption. These surface meshes can also serve as the basis for interactive collision mesh models in physics simulators [24]. It is worth noting that dense point cloud-to-surface mesh reconstruction algorithms are generally more resource-intensive compared to those based on sparse point clouds [25]. The monocular approaches using sparse or semi-dense point clouds obtained from SLAM algorithms. The obtained point cloud representation used by CARV provide a real-time solution that is power-efficient and versatile enough to function both indoors and outdoors. However, this involves a trade-off in terms of precision [19].

## B. Immersive Display

Teleoperation enables human operators to navigate robots from a remote location by using a media stream to create a mental model of the remote environment. In earlier systems, technological limitations such as the unavailability of depth cameras and constraints on network communication speeds restricted teleoperation interfaces to 2D displays [26].

However, with the availability of commodity depth cameras like the Kinect camera, researchers have developed real-time 3D reconstruction algorithms. The efficient data structures help 3D reconstructed scenes to transfer and scale to larger scenes for immersive displays like virtual reality (VR) and extended reality (XR) for robot teleoperation [27], [28]. Studies have shown that 3D immersive displays help operators increase spatial awareness and reduce the mental load by creating a more accurate mental model of the remote scene, compared to 2D displays which suffer from geometric information loss [15], [29].

As a result, the success rate of remote tasks has been found to increase with the use of 3D stereographic VR displays and the task completion time decreases [30]. However, The VR displays can also cause simulator sickness, which can negatively affect the operator's experience [31]. To address this issue, thoughtful system design and a greater degree of immersion can help reduce the occurrence of simulator sickness [30].

## C. Physics Simulator

The use of physics simulators has gained popularity in testing robots under various conditions as it is a faster and cheaper alternative to physical settings. Nonetheless, choosing the appropriate physics simulator for a specific robotics task can be challenging. The physics simulators have differing degrees of accuracy with different robot categories and real-life scenarios.

To address this issue, researchers often compare the performance of different simulators for specific tasks and robot types. The Gazebo simulator is the most frequently cited and researched due to its support for commonly used sensors and SDF meshes, as well as its compatibility with ROS1/ROS2 and multiple physics engines. However, Gazebo also has some limitations compared to other simulators, especially in terms of rendering quality [32]. The physics simulators have been used to integrate digital twins to visualize important information about the follower robot, such as force, velocity and trajectories for robot planning [33].

Another importance of physics simulators in robot teleoperation is getting contact and friction information for collision detection and avoidance using digital twins [34]. However, the process of collision detection in physics simulators is computationally demanding, and its complexity rises with the number of mesh faces present in the virtual environment [35].

## III. SYSTEM

Our operator site system operates on a single desktop computer with an Intel i9-12900K CPU and 128 GB of memory. It incorporates monocular semi-dense CARV for surface reconstruction, a Gazebo physics simulator to create an interactive virtual world, and video streaming from a Gazebo wide-angle camera sensor for head-mounted display.

## A. Surface Reconstruction

We use a real-time semi-dense monocular CARV for the incremental surface mesh reconstruction of the remote environment, harnessing the capabilities of ORB-SLAM2 for pose estimation and sparse point cloud processing [36]. Semi-dense points are created based on neighboring 50 keyframes to make it compatible with real-time performance. The semi-dense module significantly enhances precision by incorporating line and plane data, effectively eliminating outlier points and simplifying the surface mesh.

Beyond its role in providing 3D geometric information, the incremental surface mesh reconstruction also serves a crucial purpose in our system. It facilitates predictive visualization through the use of view-dependent texturing, enhancing the operator's experience with predictive visual feedback. Furthermore, it contributes to predictive haptic interaction for robot control, making it a versatile component that enhances both visualization and control aspects of our system.

As the SLAM algorithm registers new keyframes or adjusts with the bundle adjustment, the surface model is incrementally updated. Subsequently, each update results in the export of the surface mesh to an OBJ file, ensuring compatibility with the Gazebo physics simulator.

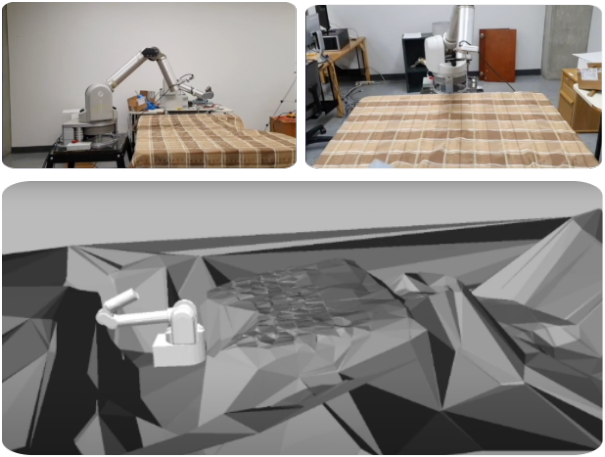

Fig. 3: Monocular surface mesh reconstruction using semi-dense CARV of the remote environment.

This compatibility enables the creation of virtual mesh models within the simulation, while keyframe and texture coordinate data pertaining to 3D rendered vertices are saved in smaller OBJ and image files for further utility and analysis. The purpose of this design is to quickly update the textured mesh. The current texture model is selected based on the specified camera pose, which is obtained from either the latest pose generated by the SLAM algorithm or the predictive texture created using the Gazebo camera pose updated by the operator's head pose.

### B. Immersive Display

The immersive display is designed to provide predictive visuals based on the operator's motion, unlike 2D displays. This immersive display not only improves spatial awareness but also reduces the mental load caused by delays.

In our system, a virtual stereo wide-angle camera within the Gazebo simulation environment serves as the operator's "eyes." To transmit this viewpoint to the operator, we employed an asynchronous ROS2 client, creating a dedicated topic for video streaming. Real-time video streaming via WebRTC was facilitated using the open-source Python library *aiortc*.

For seamless integration with various VR headsets, we used the A-frame framework, which leverages Three.js for rendering virtual and augmented reality applications in web browsers. This approach ensures compatibility across different VR headsets and simplifies implementation.

The operator's head pose, obtained through the framework, controls the Gazebo camera using the Gazebo ROS1 plugin. Pose data is transmitted via the data channel, alongside the video channel created with WebRTC. This allows the operator to control the camera pose and head motion. To enable camera control without requiring physical movement, the hand controller joysticks are integrated with the operator's pose to update the Gazebo camera's rendering.

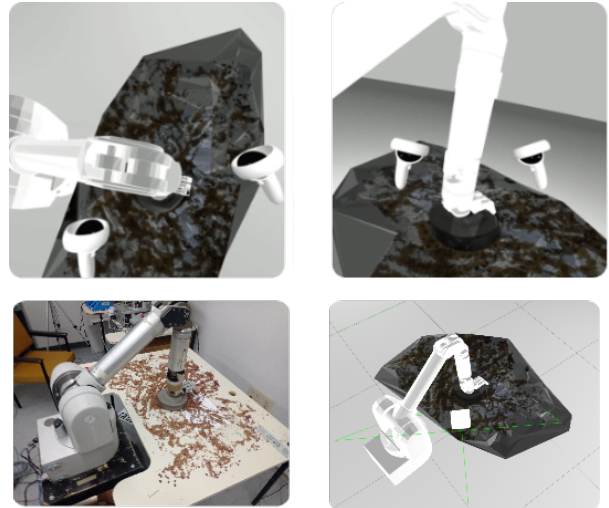

Fig. 4: **(Top)** Corresponding camera feed in the VR is based on the updated camera pose, **(Bottom Left)** The real world scene depicting a mine field, **(Bottom Right)** Wide angle camera as cube controlled by the operator to change view pose.

### C. Physics Simulator: Gazebo

The physics simulator serves multiple critical purposes in our system. It not only seamlessly integrates the updated surface mesh model of the remote environment and the digital twin of the follower robot but also plays a pivotal role in enhancing operator experience. Specifically, it provides predictive haptic feedback by utilizing interactions between the surface mesh model and the digital twin of the follower robot. Additionally, it offers predictive visualization based on operator motion using camera sensors, effectively eliminating delay problems for both visualization and control.

We chose Gazebo for its seamless integration of the updated surface mesh of the remote environment and the WAM Robot's digital twin. To expand Gazebo's capabilities and model compatibility, we used Gazebo plugins. These plugins enable dynamic movement within the simulation environment by capturing joint positions, velocities, and efforts for the digital twin. Communication between the remote WAM arm robot and the digital twin is facilitated through ROS topics published by libbarrett to the Gazebo model plugin.

Interactions between the digital twin and the calibrated surface collision mesh are monitored for contact detection, which includes identifying contact normals. Subsequently, the ODE physics engine calculates Force/Torque magnitudes and friction values for these interactions.

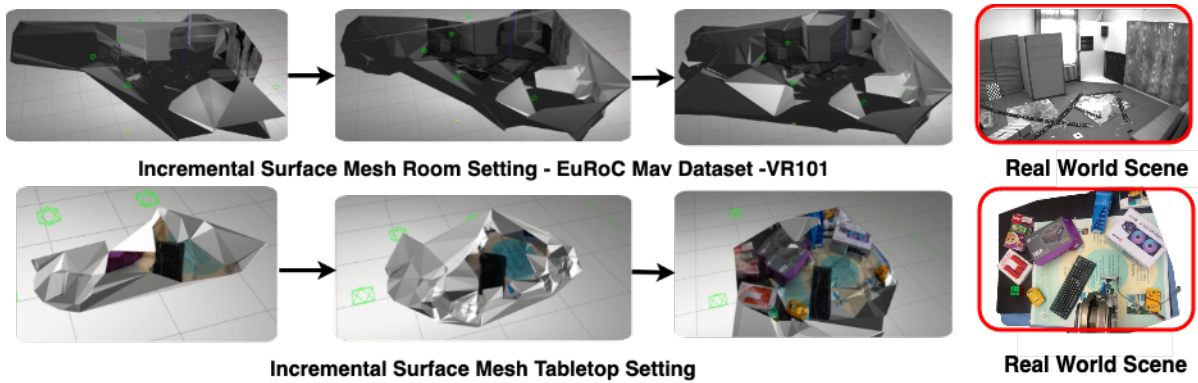

Fig. 5: Incremental surface mesh being updated in physics simulator in EuRoC MAV Dataset Vicon Room 101(**VR101**). **(Top)** with the actual scene **(Top Right)** and incremental surface mesh being updated in physics simulator for the real life tabletop scenario **(Bottom)** with the actual scene **(Bottom Right)**.

We used a digital twin of the WAM arm robot model based on Institut de Robòtica's URDF WAM Models and STL files. The model files are later converted to Gazebo SDF files by Benjamin Blumer. The model files that are compatible with Gazebo 11 version can be found here. The digital twin of the WAM arm robot is used to visualize up to 17 joints of the WAM arm robot.

## IV. EXPERIMENTS

### A. Surface Reconstruction

In our system, real-time surface mesh reconstruction is achieved using an Intel RealSense D435 global shutter camera, which provides 1280x720 RGB frames. Unlike rolling shutter webcams, the D435's global shutter is compatible for SLAM systems. This setup operates at 15.46 FPS on an Intel i9-12900K processor. The resulting surface meshes are depicted in the lower section of Figure 5.

We compared three monocular surface mesh reconstruction methods that use sparse point cloud and Delaunay triangulation to create a surface mesh of the scene [10], [19]. We compared the algorithms based on precision and completeness metrics using EuRoC MAV Vicon Room 101 (VR101) [37] benchmark. Our algorithm improves precision and completeness while reducing the vertices and mesh faces (Table I).

| Method | Vertices | Faces | Precision | Completeness |
|---|---|---|---|---|
| Lovi *et al.* [10] | 531142 | 148582 | 74.5% | 70.98% |
| He *et al.* [19] | 50666 | 32012 | 88.33% | 77.21% |
| Ours | **31986** | **22998** | **96.8%** | **88.62%** |

TABLE I: Comparison between the surface mesh of **VR101** room setting created by different CARV approaches on completeness and precision.

### B. Immersive Display

The video stream from the Gazebo camera sensor was sent to the VR headset, an Oculus Quest-2 with hand controllers, offering a high-resolution display. Our system supports video streaming over the local network at 1696x1600 resolution, running at 30 FPS with approximately 10ms latency.

### C. Physics Simulator

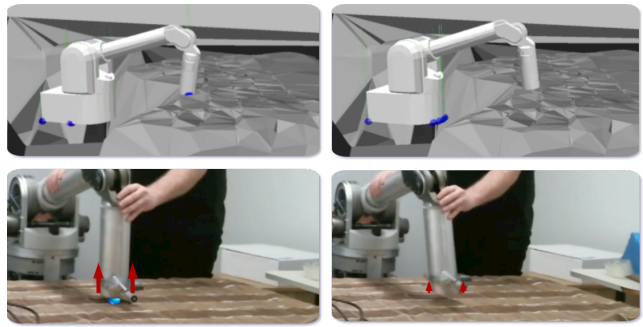

Fig. 6: **Top :** The contact between the surface mesh and the WAM digital twin robot was calculated for haptic feedback before 20 mm of the physical contact using *min_depth* parameter. **Bottom :** Haptic feedback to WAM arm robot arm by surface collision mesh from physics simulator (**Blue Sphere**) visualizing contact between the table and the robot arm end effector, (**Red Arrow**) representing the contact forces calculated using surface mesh face normals.

In our experiments, we configured a Gazebo world to replicate the remote environment in real-time. Real-time model updates are achieved using Gazebo ROS plugins. Initially, communication between the digital twin of the WAM arm robot and the physical WAM arm robot relied on the Gazebo ROS plugin and *libbarrett*, operating at a rate of 30 frames per second (FPS). However, this update rate was insufficient for providing haptic feedback while moving the WAM arm robot, leading to oscillations due to high latency.

To address this issue, we developed a custom Gazebo model plugin to internally integrate *libbarrett* ROS topics and services with Gazebo. This plugin utilizes *libbarrett* to retrieve joint data and supply haptic feedback with 250 frames per second. Communication between the digital twin and the physical WAM arm robot is carried out via ROS1.

For accurate force and torque values, we employed Gazebo force sensors. However, due to occasional inaccuracies, we opted to use constant force values defined in the plugin to ensure consistent haptic feedback. To enhance the safety of the WAM arm robot, we enabled *collision_without_contact* in the Gazebo physics simulator configuration, allowing us to detect contact before collisions without obtaining contact force values.

Collision detection can be computationally intensive, impacting real-time performance as the digital twin's complexity increases. Nevertheless, our Gazebo world configuration optimizes collision detection by disabling wind, atmosphere, and shadows from renderings. We further improved real-time performance by adjusting update rates and step sizes to maximize CPU utilization. Table II provides a comparison of results with and without optimization for various surface reconstruction algorithms.

By default, Gazebo disables projection rendering for its camera sensor, limiting visualizations to the program's GUI. However, we modified the source code to enable rendering for Gazebo camera sensors, harnessing the open-source nature of Gazebo. This modification allows the digital twin to provide visualization of contact, inertia, and efforts to the immersive display through the Gazebo camera sensor.

In our experiments, we compared the performance of the Gazebo ROS2 and ROS1 camera plugins. The ROS2 plugin achieved 30 FPS at 4K resolution, while the ROS1 plugin reached 30 FPS at 640x480 resolution for streaming. To capture necessary information from the mesh model and digital twin, we configured the Gazebo camera sensor with a 1.57-radian horizontal field of view. An important finding was that scene illumination improved camera projection, prompting us to incorporate a directional light sensor based on the camera's pose, which was updated at 30 FPS by the human operator via the Gazebo ROS1 plugin service.

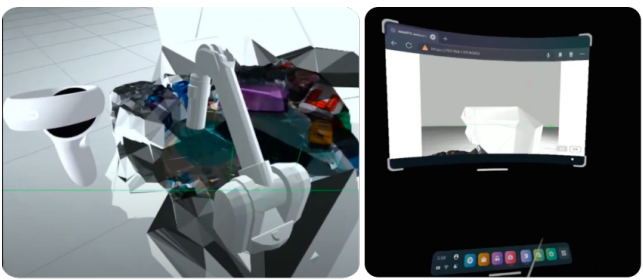

Fig. 7: Predictive texture and surface mesh on immersive display streamed by ROS2 from Gazebo wide-angle camera.

The Gazebo surface mesh model is updated using ROS1 plugins. For efficient real-time updates, we employed two distinct mesh models: one for the surface mesh of all scanned scenes and another for the textured mesh of a single frame. In our experiments, the room-scale surface mesh OBJ file occupied approximately 1.4 MB of disk space, while the textured mesh OBJ file was 300-600 KB, with an additional 300 KB JPEG image frame at 1280x720 resolution.

| Method | Memory(MB) | RTF w/o opt | RTF w opt |
|---|---|---|---|
| Lovi *et al.* [10] | 18.7 | 0.11 | 0.29 |
| He *et al.* [19] | 2.1 | 0.225 | 0.471 |
| Ours | **1.4** | **0.255** | **0.52** |

TABLE II: Comparison between the surface mesh of the from Euroc Vicon Room 101 benchmarks (**VR101**) [37] setting physics simulator real-time factor (**RTF**) with w/wo optimization, (RTF being 1.00 to be the optimal RTF.)

In the table-top setting, both the surface mesh OBJ file and the textured mesh had a disk size around 900 KB. Model updates and additions to the simulation environment were facilitated using the *spawn_model* ROS1 service, completing the process in 60 ms for a 1.4 MB OBJ file. Removing models from the Gazebo simulation environment was achieved using the *delete_model* Gazebo ROS1 plugin service, taking 250 ms with *rospy*.

## V. CONCLUSION

We created this novel predictive system unlike any other systems to directly deliver haptic feedback in response to the operator's motions,, eliminating any delays associated with the haptic follower robot.

Accurate virtual models could help the operator navigate and manipulate tasks by minimizing the mental load of complex scene models and estimating the robot's state in the remote scene. To create a virtual model of the remote scene, we utilized an incremental monocular semi-dense ORBSLAM-2 based surface mesh reconstruction model. With the integration of the digital twin WAM arm robot, the virtual intractable representation of the remote environment is completed to provide predictive visual and haptic feedback. Furthermore, the immersive VR displays help the operator change the view independently from the remote scene view, enhancing flexibility without increasing bandwidth requirements. Our proposed teleoperation system offers a promising solution to address the challenges associated with robot teleoperation and enables effective navigation and manipulation in complex 3D environments in real-time with mobility limitations.